\documentclass[11pt,a4paper]{article}
\usepackage[hyperref]{DomainAdaptationMaerz}
\usepackage{times}
\usepackage{latexsym}
\usepackage{graphicx}

\usepackage{url}

\usepackage{pgfplots}
\pgfplotsset{compat=1.14}
\pgfplotsset{ 
  compat=1.12, 
  /pgf/number format/use comma 
}

\aclfinalcopy 
\def\aclpaperid{1107} 

\title{Domain adaptation for part-of-speech tagging of noisy user-generated text}

\author{Luisa M\"arz \and Dietrich Trautmann \and Benjamin Roth \\
  CIS, University of Munich (LMU) Munich, Germany  \\
 {\tt \{luisa.maerz, dietrich, beroth\}@cis.lmu.de} 
}

\date{}

\begin{document}
\maketitle
\begin{abstract}
The performance of a Part-of-speech (POS) tagger is highly dependent on the domain of the processed text, and for many domains there is no or only very little training data available.
This work addresses the problem of POS tagging noisy user-generated text using a neural network. 
We propose an architecture that trains an out-of-domain model on a large newswire corpus, 
and transfers those weights by using them as a prior for a model trained on the target domain (a data-set of German Tweets) for which there is very little annotations available.
The neural network has two standard bidirectional LSTMs at its core.
However, we find it crucial to also encode a set of task-specific features, and to obtain reliable (source-domain and target-domain) word representations.
Experiments with different regularization
techniques such as early stopping, dropout and fine-tuning the domain adaptation prior weights are conducted. 
Our best model uses external weights from the out-of-domain model, as well as feature embeddings, pretrained word and sub-word embeddings and achieves a tagging accuracy of slightly over 90\%, improving on the previous state of the art for this task.
\end{abstract}


\section{Introduction}
Part-of-speech (POS) tagging is a prerequisite for many applications
and necessary for a wide range of tools for computational
linguists.
The state-of-the art method to implement a tagger is to use neural networks \citep{hovy,yang2018design}.
The performance of a POS tagger is highly dependent on the domain of the processed text and the availability of sufficient training data \citep{schnabel2014flors}. Existing POS taggers for canonical German text
already achieve very good results around 97\% accuracy, e.g. \citep{schmid,DBLP:journals/corr/PlankSG16}.
When applying these trained models to out-of-domain data the performance decreases drastically.

One of the domains where there is not enough data is online conversational text in platforms such as Twitter, where the very informal language exhibits many phenomena that differ significantly from canonical written language.

In this work, we propose a neural network that combines a character-based encoder and embeddings of features from previous non-neural approaches (that can be interpreted as an inductive bias to guide the learning task).
We further show that the performance of this already effective tagger can be improved significantly by incorporating external weights using a mechanism of domain-specific L2-regularization during the training on in-domain data. 
This approach establishes state-of-the-art results of 90.3\% accuracy on the German Twitter corpus of \citet{rehbein}. 

\section{Related Work}
The first POS tagging approach for German Twitter data was conducted by \citet{rehbein} and reaches an accuracy of 88.8\% on the test set using a CRF. They use a feature set with eleven different features and an extended version of the STTS \citep{stts} as a tagset.
\citet{Gimpel:2011:PTT:2002736.2002747} developed a tagset for English Twitter data and report results of 89.37\% on their test set using a CRF with different features as well.
POS tagging for different languages using a neural architecture was successfully applied by \citet{DBLP:journals/corr/PlankSG16}. The data comes from the Universal Dependencies project\footnote{\url{http://universaldependencies.org}} and mainly contains German newspaper texts and Wikipedia articles.

The work of \citet{barone} investigates different regularization mechanisms in the field of domain adaptation. They use the same L2 regularization mechanism for neural machine translation, as we do for POS tagging.

\section{Data}

\subsection{Tagset}
The Stuttgart-T\"ubingen-TagSet (STTS, \citet{stts}) is widely used as the state-of-the-art tagset for POS tagging of German.
\citet{barzt} show that the STTS is not sufficient when working with textual data from online social platforms, as online texts do not have the same characteristics as formal-style texts, nor are identical to spoken language.
Online conversational text often contains contracted forms, graphic reproductions of spoken language such as prolongations, interjections and grammatical inconsistencies as well as a high rate of misspellings, omission of words etc.

For POS tagging we use the tagset of \citet{rehbein}, where (following \citet{Gimpel:2011:PTT:2002736.2002747}) additional tags are provided to capture peculiarities of the Twitter corpus.
This tagset provides tags for @-mentions, hashtags and URLs.
They also provide a tag for non-verbal comments such as \textbf{*Trommelwirbel*} (drum-roll).
Additional, complex tags for amalgamated word forms were
used (see \citet{Gimpel:2011:PTT:2002736.2002747}).
Overall the tagset used in our target domain contains 15 tags more than the original STTS.

\subsection{Corpora}
Two corpora with different domains are used in this work. One of them is the TIGER corpus and the other is a collection of German Twitter data.

The texts in the TIGER corpus \citep{brants} are taken from the Frankfurter Rundschau newspaper and date from 1995 over a period of two weeks. The annotation of the corpus was created semi automatically.
The basis for the annotation of POS tags is the STTS.
The TIGER corpus is one of the standard corpora for German in NLP and contains 888.505 tokens.

The Twitter data was collected by \citet{rehbein} within eight months in 2012 and 2013.
The complete collection includes 12.782.097 distinct tweets, from which 1.426 tweets were randomly selected for manual annotation with POS tags.
The training set is comparably small and holds 420 tweets, whereas the development and test set hold around 500 tweets each (overall 20.877 tokens). 
Since this is the only available German annotated Twitter corpus, we use it for this work.

\subsection{Pretrained word vectors}
The usage of pretrained word embeddings can be seen as a standard procedure in NLP to improve the results with neural networks (see \citet{hovy}. 

\subsection{FastText}
FastText\footnote{\url{https://fasttext.cc}} provides pretrained sub-word embeddings for 158 different languages and allows to obtain word vectors for out-of-vocabulary words. The pretrained vectors for German are based on Wikipedia articles and data from Common Crawl\footnote{\url{https://commoncrawl.org}}.
We obtain 97.988 different embeddings for the tokens in TIGER and the Twitter corpus of which 75.819 were already contained in Common Crawl and 22.171 were inferred from sub-word units.

\subsection{Word2Vec}
Spinningbytes\footnote{\url{https://www.spinningbytes.com}} is a platform for different applications in NLP and provides several solutions and resources for research.
They provide word embeddings for different text types and languages, 
 including Word2Vec \cite{mikolov2013efficient} vectors pretrained on 200 million German Tweets. 
Overall 17.030 word embeddings form the Spinningbytes vectors are used (other words are initialized all-zero).

\subsection{Character level encoder}
\citet{lample} show that the usage of a character level encoder is expedient when using bidirectional LSTMs. Our implementation of this encoder follows Hiroki Nakayama (2017)\footnote{\url{https://github.com/Hironsan/anago}}, where character embeddings are passed to a bidirectional LSTM and the output is concatenated to the word embeddings.

\begin{figure}[h]
\begin{center}
\includegraphics[width=230pt]{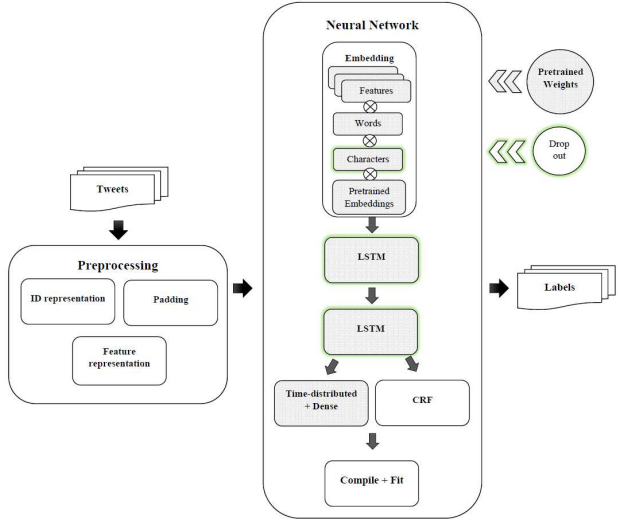}
\caption{Final architecture of the neural model. Layers that are passed pretrained weights are hatched in gray. Dropout affected layers are highlighted in green.}
\label{final-graph}
\end{center}
\end{figure}

\section{Experiments}
This section describes the proposed architecture of the neural network and the conditional random field used in the experiments.
For comparison of the results we also experiment with jointly training on a merged training set, which contains the Twitter and the TIGER training sets.

\subsection{Methods}

\subsubsection{Conditional random field baseline}
The baseline CRF of \citet{rehbein} achieves an accuracy of 82.49\%. To be comparable with their work we implement a CRF equivalent to their baseline model.
Each word in the data is represented by a feature dictionary. We use the same features as \citeauthor{rehbein} proposed for the classification of each word. These are the lowercased word form,  word length, number of uppercase letters, number of digits and occurrence of a hashtag, URL, @-mention or symbol.

\subsubsection{Neural network baseline}
The first layer in the model is an embedding layer. 
The next layers are two bidirectional LSTMs. 
The baseline model uses softmax for each position in the final layer and is optimized using Adam core with a learning rate of 0.001 and the categorical crossentropy as the loss function.

\subsubsection{Extensions of the neural network}
The non neural CRF model benefits from different features extracted from the data. Those features are not explicitely modeled in the neural baseline model, and we apply a feature function for the extended neural network.
We include the features used in the non-neural CRF for hashtags and @-mentions. 
In addition, we capture orthographic features, e.g., whether a word starts with a digit or an upper case letter. 
Typically, manually defined features like these are not used in neural networks, as a neural network should take over feature engineering completely. Since this does not work optimally, especially for smaller data sets, we have decided to give the neural network this type of information as well. Thus we combine the advantages of classical feature engineering and neural networks. This also goes along with the observations of \citet{DBLP:journals/corr/abs-1811-08757} and \citet{W17-6304}, who both show that adding conventional lexical information improves the performance of a neural POS tagger.
All words are represented by their features and for each feature type an embedding layer is set up within the neural network in order to learn vectors for the different feature expressions. Afterwards all the feature embeddings are added together. As the next step we use the character level layer mentioned in section 3.6 \citep{lample}. The following vector sequences are concatenated at each position and form the input to the bidirectional LSTMs:
\begin{itemize}
\setlength{\parskip}{1pt}
\item Feature embedding vector
\item character-level encoder
\item FastText vectors
\item Word2Vec vectors
\end{itemize}

\subsubsection{Domain Adaptation and regularization}
We train the model with the optimal setting on the TIGER corpus, i.e., we prepare the TIGER data just like the Twitter data and extract features, include a character level layer and use pretrained embeddings.
We extract the weights $\hat W$ that were optimized with TIGER.
The prior weights $\hat W$ are used during optimization as a regularizer for the weights $W$ used in the final model (trained on the Twitter data).
This is achieved by adding the penalty term $R_W$,  as shown in Equation \ref{eq:regularization}, to the objective function (cross-entropy loss).
\begin{equation}
\label{eq:regularization}
R_W = \lambda ||W - \hat{W} ||_2^2
\end{equation}
The regularization is applied to the weights of the two LSTMs, the character LSTM, to all of the embedding layers and to the output layer.

As a second regularization mechanism we include dropout for the forward and the backward LSTM layers. 
We also add 1 to the bias of the forget gate at initialization, since this is recommended in \citet{Jozefowicz:2015:EER:3045118.3045367}.
Additionally, we use early stopping.
Since the usage of different regularization techniques worked well in the experiments of \citet{barone}, we also tried the combination of different regularizers in this work. Figure \ref{final-graph} shows the final architecture of our model.

\subsection{NCRF++}
We also report results obtained by training the sequence labelling tagger of \citet{yang2018ncrf}, NCRF++. They showed that their architecture produces state-of-the-art models across a wide range of data sets \citep{yang2018design} so we used this standardized framework to compare it with our model.

\section{Results}
\subsection{Experimental Results}
Table \ref{testresult} shows the results on the Twitter test set.
The feature-based baseline CRF outperforms the baseline of the neural net with more than 20 percentage points. 
After adding the feature information, the performance of the neural baseline is improved by 13 percentage points, which is understandable, because many German POS tags are case sensitive. 

\begin{table}[ht]
\centering
\begin{tabular}{|l|l|} 
\hline
\textbf{experiment} & \textbf{accuracy} \\ \hline \hline
\textit{baseline crf} & 0.831 \\ \hline
\textit{baseline neural model} & 0.634 \\ \hline
\textit{neural model} & \\
\hspace{0.4cm} \textit{+features} & 0.768 \\
\hspace{0.4cm} \textit{+character embeddings} & 0.796 \\ 
\hspace{0.4cm} \textit{+pretrained word vectors} & 0.845 \\
\hspace{0.4cm} \textit{+l2 domain adaptation} & 0.896\\
\hspace{0.4cm} \textit{+dropout} & \textbf{0.903}\\ \hline
\textit{neural model joint training} & 0.894 \\ \hline
\textit{final CRF of Rehbein 2013 } & 0.888 \\ \hline
\textit{NCRF++  system} & 0.887 \\ \hline
\end{tabular}
\caption{Results on the test set using the time-distributed layer. 
}
\label{testresult}
\end{table}

The model's performance increases by another 3 percentage points if the character level layer is used.
Including the pretrained embeddings, FastText and Word2Vec vectors, the accuracy is 84.5\%, which outperforms the CRF baseline. 

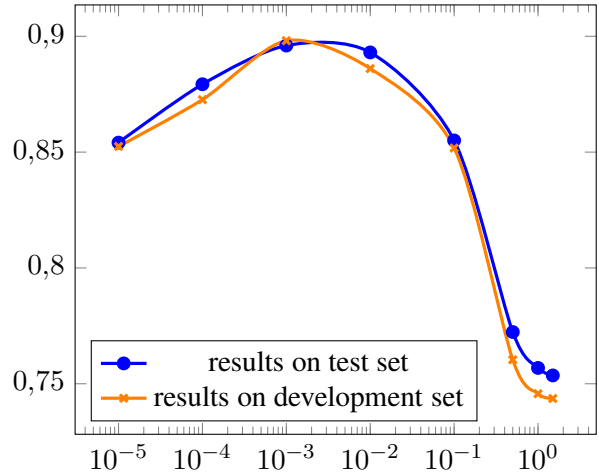
\begin{figure}
    \centering
\begin{tikzpicture}
\begin{semilogxaxis}[legend pos=south west, 
    ]
\addplot[smooth,mark=*,blue, very thick]
coordinates {
(0, 0.840495756)
(0.00001, 0.854102115)
(0.0001, 0.879294086)
(0.001, 0.896)
(0.01, 0.893035161)
(0.1, 0.85504513)
(0.5, 0.772329247)
(1, 0.756836858)
(1.50, 0.753603664)
};
\addlegendentry{results on test set}
\addplot[smooth,color=orange,mark=x, very thick]
coordinates {
(0,	0.838910398)
(0.00001, 0.852461283)
(0.0001, 0.872649336)
(0.001, 0.898091814)
(0.01, 0.886061947)
(0.1, 0.851769912)
(0.5, 0.760370575)
(1, 0.745713496)
(1.50, 0.743639381)
};
\addlegendentry{results on development set}
\end{semilogxaxis}
\end{tikzpicture}
    \caption{Influence of fine-tuning on the results on dev and test set in accuracy (y-axis). 
The x-axis corresponds to the different $\lambda$ values.}
    \label{l2norm}
\end{figure}

Figure \ref{l2norm} shows the impact of domain adaptation and fine-tuning the prior weight.
The value of the $\lambda$ parameter in the regularization formula \ref{eq:regularization} can control the degree of impact of the weights on the training. 
Excluding the pretrained weights means that $\lambda$ is 0.
We observe an optimal benefit from the out-of-domain weights by using a $\lambda$ value of 0.001. This is in line with the observations of \citet{barone} for transfer-learning for machine translation.\\
\indent Overall the addition of the L2 fine-tuning can improve the tagging outcome by 5 percentage points, compared to not doing domain adaptation. A binomial test shows that this improvement is significant.
This result confirms the intuition that the tagger can benefit from the pretrained weights.
On top of fine-tuning different dropout rates were added to both directions of the LSTMs for the character level layer and the joint embeddings. 
A dropout rate of 75\% is optimal in our scenario, and it increases the accuracy by 0.7 percentage points.\\
\indent The final 90.3\% on the test set outperform the results of \citet{rehbein} by 1.5 percentage points.
Our best score also outperforms the accuracy obtained with the NCRF++ model. This shows that for classifying noisy user-generated text, explicit feature engineering is beneficial, and that the usage of domain adaptation is expedient in this context.
Joint training, using all data (out-of-domain and target domain), can obtain an accuracy score of 89.4\%, which is about 1 percentage point worse than using the same data with domain adaptation. The training setup for the joint training is the same as for the other experiments and includes all extensions except for the domain adaptation. \\

\vspace*{-0.5cm}

\begin{figure}
\centering
\includegraphics[width=225pt]{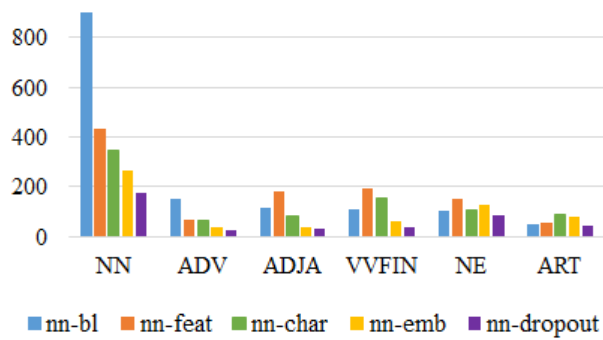}
    \caption{Total number of errors for the six most frequent POS-tags and different experimental settings}
\label{errortypes}
\end{figure}

\subsection{Error Analysis}
The most frequent error types in all our systems were nouns, proper nouns, articles, verbs, adjectives and adverbs as pictured in figure \ref{errortypes}. 
By including the features the number of errors can be reduced drastically for nouns.
Since we included a feature that captures upper
and lower case, and nouns as well as proper nouns are written upper case in German, the model can benefit from that information. 
The pretrained word embeddings also help classifying nouns, articles, verbs, adjectives and adverbs. Only the errors with proper nouns increase slightly. 
Compared to only including the features, the model can benefit from adding both, the character level layer and the pretrained word vectors, while the results for tagging proper nouns and articles are still slightly worse than the baseline.
In contrast the final experimental setup can optimize the results for every POS tag compared to the baseline, see figure \ref{errortypes}. Slightly in case of articles and proper nouns, but markedly for the other tags.
A comparison of the baseline errors and the errors of the final system shows that Twitter specific errors, e.g. with @-mentions or URLs, can be reduced drastically. Only hashtags still pose a challenge for the tagger. In the gold standard words with hashtags are not always tagged as such, but sometimes are classified as proper nouns. This
is due to the fact that the function of the token in the sentence is the one of a proper noun. Thus the tagger has decision problems with these hashtags.
Other types of errors, such as confusion of articles or nouns, are not Twitter-specific issues, but are often a problem with POS tagging and can only be fixed by general improvement of the tagger.

\section{Conclusion}
We present a deep learning based fine-grained POS tagger for German Twitter data using both domain adaptation and regularization techniques. 
On top of an efficient POS tagger we implemented domain adaptation by using a L2-norm regularization mechanism, which improved the model's performance by 5 percentage points. Since this performance is significant we conclude that fine-tuning and domain adaptation techniques can successfully be used to improve the performance when training on a small target-domain corpus.

Our experiments show that the combination of different regularization techniques is recommendable and can further optimize already efficient systems.

The advantage of our approach is that we do not need a large
annotated target-domain corpus, but only pretrained weights. 
Using a pretrained model as a prior for training on a small amount of data is done within minutes and therefore very practicable in real world scenarios.

\bibliography{DomainAdaptationMaerz}
\bibliographystyle{DomainAdaptationMaerz}




\end{document}